\documentclass[10pt,twocolumn,letterpaper]{article}

\usepackage[pagenumbers]{cvpr} 

\definecolor{cvprblue}{rgb}{0.21,0.49,0.74}
\usepackage[pagebackref,breaklinks,colorlinks,allcolors=cvprblue]{hyperref}
\usepackage{wrapfig}
\usepackage{multirow}
\usepackage{colortbl}
\usepackage[dvipsnames]{xcolor}

\definecolor{mygray}{gray}{0.9}

\begin{document}

\title{Inversion Circle Interpolation: \\ Diffusion-based Image Augmentation for Data-scarce Classification}

\author{Yanghao Wang, Long Chen\\
The Hong Kong University of Science and Technology\\
{\tt\small \href{https://github.com/scuwyh2000/Diff-II}{https://github.com/scuwyh2000/Diff-II}}
}

\twocolumn[{%
    \maketitle
    \begin{center}
        \centering
        \vspace{-1em}
        \centering
\includegraphics[width=1\linewidth]{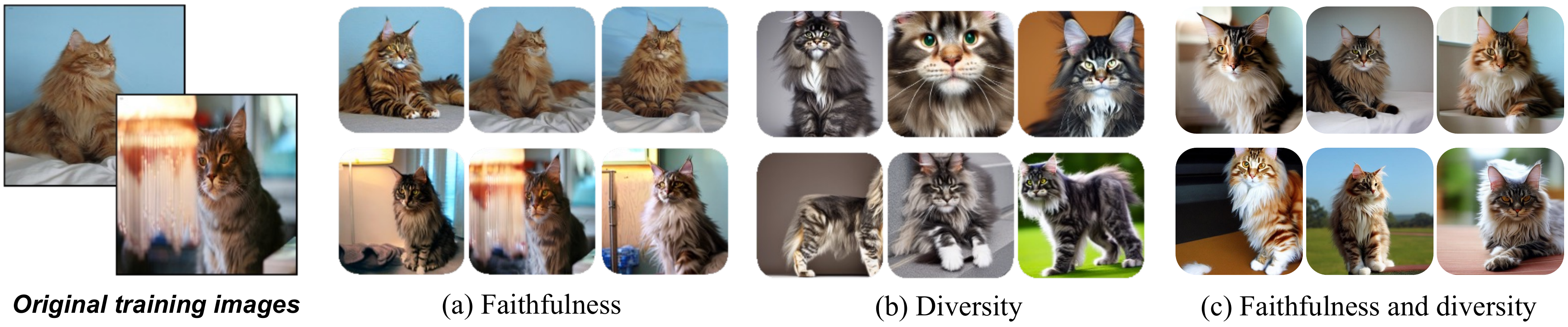}
\vspace{-2em}
\captionof{figure}{Given training images, data augmentation aims to generate new \emph{faithful} and \emph{diverse} synthetic images. (a) These synthetic images are faithful but not diverse. (b) These synthetic images are diverse but not faithful. (c) These synthetic images are both faithful and diverse.}
\label{fig:1}
    \end{center}%
}]

\begin{abstract}
Data Augmentation (DA), \ie, synthesizing faithful and diverse samples to expand the original training set, is a prevalent and effective strategy to improve the performance of various data-scarce tasks. With the powerful image generation ability, diffusion-based DA has shown strong performance gains on different image classification benchmarks. In this paper, we analyze today's diffusion-based DA methods, and argue that they cannot take account of both \emph{faithfulness} and \emph{diversity}, which are two critical keys for generating high-quality samples and boosting classification performance. To this end, we propose a novel Diffusion-based DA method: \textbf{Diff-II}. Specifically, it consists of three steps: 1) \emph{Category concepts learning}: Learning concept embeddings for each category. 2) \emph{Inversion interpolation}: Calculating the inversion for each image, and conducting circle interpolation for two randomly sampled inversions from the same category. 3) \emph{Two-stage denoising}: Using different prompts to generate synthesized images in a coarse-to-fine manner. Extensive experiments on various data-scarce image classification tasks (\eg, few-shot, long-tailed, and out-of-distribution classification) have demonstrated its effectiveness over state-of-the-art diffusion-based DA methods.


\end{abstract}    
\section{Introduction}
\label{sec:1}
Today's visual recognition models can even outperform us humans with sufficient training samples. However, in many different real-world scenarios, it is not easy to collect adequate training data for some categories (\ie, data-scarce scenarios). For example, since the occurrence frequency of various categories in nature follows a long-tailed distribution, there are many rare categories with only limited samples~\citep{o2004kendall,van2017devil}. To mitigate this data scarcity issue, a prevalent and effective solution is \textbf{Data Augmentation (DA)}. Based on an original training set with limited samples, DA aims to generate more synthetic samples to expand the training set.

For DA methods, there are two critical indexes: \emph{faithfulness} and \emph{diversity}~\citep{sajjadi2018assessing}. They can not only show the quality of synthesized samples, but also influence the final classification performance. More specifically, \textbf{faithfulness} indicates that the synthetic samples need to retain the characteristics of the corresponding category (\cf, Figure~\ref{fig:1} (a)), \ie, \emph{the faithfulness confirms that the model learns from correct category knowledge}. \textbf{Diversity} indicates that the synthetic samples should have different contexts from the original training set and each other (\cf, Figure~\ref{fig:1} (b)), \ie, \emph{the diversity ensures that the model learns the invariable characteristics of the category by seeing diverse samples}.

\begin{figure*}[t]
    \centering
    \includegraphics[width=1\linewidth]{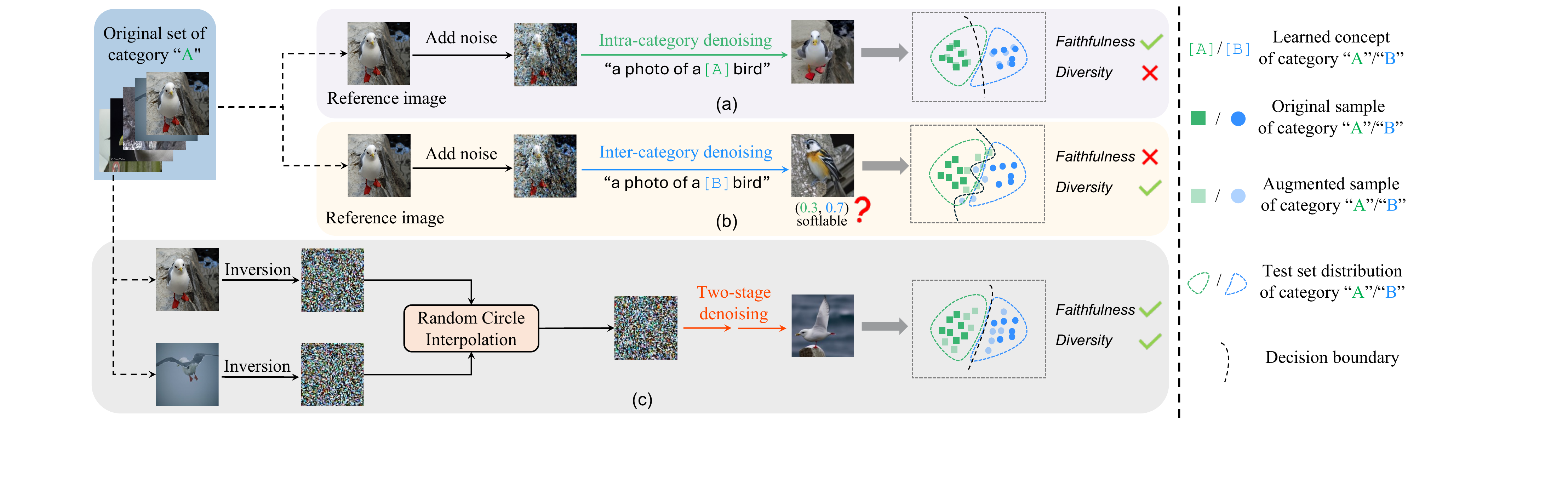}
    \vspace{-2em}
    \caption{a) \textbf{Intra-category DA}: Given a reference image (from the original set), it adds some noise and denoises with a prompt containing the same category concept (\eg, concept ``\texttt{[A]}'' for category A image). (b) \textbf{Inter-category DA}: Different from Intra-category DA, it denoises with a prompt containing a different category concept (\eg, concept ``\texttt{[B]}'' for category A image). (c) \textbf{Ours}: It first calculates the inversion for each image, and conducts random circle interpolation for two images of the same category. Then, it denoises in a two-stage manner with different prompts.}
    \label{fig:2}
\end{figure*}

With the photo-realistic image generation ability of today's diffusion models~\citep{ho2020denoising,nichol2021improved}, a surge of diffusion-based DA methods has dominated the image classification task\footnote{DA has been used in various image classification settings. In this paper, we focus on data-scarce scenarios (more discussions are in the Sec.~\ref{sec:2}).}.
Typically, diffusion-based DA methods reformulate image augmentation as an image editing task, which consists of two steps:
1) \emph{Noising Step}: They first randomly sample an image from the original training set as a reference image and then add some noise to the reference image. 2) \emph{Denoising Step}: They then gradually denoise this noisy reference image conditioned on a category-specific prompt. After the two steps, a new synthesized training image was generated. Following this framework, the pioneer diffusion-based DA work~\citep{he2022synthetic} directly uses a hand-crafted template containing the reference image's category label as the prompt (\ie, intra-category denoising). These handcrafted prompts work well on general datasets with a broad spectrum of category concepts (\eg, CIFAR-10~\citep{krizhevsky2009learning}). However, these few words (with only category name) can not guide the diffusion models to generate images with specific and detailed characteristics, especially for datasets with fine-grained categories (\eg, Stanford Cars~\citep{krause20133d}). 

To further enhance the generalization ability, subsequent diffusion-based DA methods try to improve the quality of synthesized samples from the two key characteristics. Specifically, \textbf{to improve faithfulness}, \cite{trabucco2023effective} replace category labels with more fine-grained learned category concepts. As shown in Figure~\ref{fig:2}(a), they first learn a specific embedding ``\texttt{[A]}'' for ``category A'' bird, and then replace the fixed category name with the learned concept in the prompt. These learnable prompts can somewhat preserve fine-grained details for different categories. However, the fixed combination of a reference image and its corresponding category concept always results in similar synthetic samples (\ie, limited diversity). On the other side, \textbf{to improve diversity}, \cite{wang2024enhance} use prompts containing different category concepts (\eg, ``\texttt{[B]}'') from the reference image (\ie, inter-category denoising). This operation can generate images with ``intermediate'' semantics between two different categories. However, it inherently introduces another challenging problem to obtain an ``accurate'' soft label for each synthetic image, which affects faithfulness to some extent (\cf, Figure~\ref{fig:2}(b)). Based on these above discussions, we can observe that: \emph{current state-of-the-art diffusion-based DA methods cannot take account of both faithfulness and diversity}, which results in limited improvements on the generalization ability of downstream classifiers.

In this paper, we propose a simple yet effective \underline{\textbf{Diff}}usion-based \underline{\textbf{I}}nversion \underline{\textbf{I}}nterpolation method: \textbf{Diff-II}, which can generate both faithful and diverse augmented images. As shown in Figure~\ref{fig:2}(c), Diff-II consists of three steps: 1) \emph{Category Concepts Learning}: To generate faithful images, we learn a specific embedding for each category (\eg, ``\texttt{[A]}'' for category A) by reconstructing the images of the original training set. 2) \emph{Inversion Interpolation}: To improve diversity while maintaining faithfulness, we calculate the inversion\footnote{In the image generation field, the inversion refers to a latent representation that can be used to reconstruct the corresponding original image by the generative model.} for each image of the original training set. Then, we sample two inversions from the same category and conduct interpolation. The interpolation result corresponds to a subsequent high-quality synthetic image. 3) \emph{Two-stage Denoising}: To further improve the diversity, we prepare some suffixes\footnote{More suffix examples are shown in the appendix. \label{footnote:appendix}} (\eg, ``\emph{flying over water}'', ``\emph{standing on a tree branch}'') that can summarize the high-frequency context patterns of the original training set. Then, we split the denoising process into two stages by timesteps. In the first stage, we denoise the interpolation results guided by a prompt containing the learned category concept  and a randomly sampled suffix, \eg, ``a photo of a \texttt{[A]} bird \texttt{[suffix]}.'' This design can inject perturbation into the early-timestep generation of context and finally contributes to diversity. In the second stage, we replace the prompt with ``a photo of a \texttt{[A]} bird'' to refine the character details of the category concept.

Specifically, we first utilize some parameter-efficient fine-tuning methods (\ie, low-rank adaptation~\citep{hu2021lora} and textual inversion~\citep{gal2022image}) to learn the concept embedding for each category. Then, we acquire the DDIM inversion~\citep{song2020denoising} for each image from the original set conditioned on the learned concept. After that, we randomly sample two inversions within one category as one pair and conduct interpolation with random strengths. To align the distribution of interpolation results with standard normal distribution and get a larger interpolation space, we conduct random circle interpolation. Since each pair of images used for inversion interpolation belongs to the same concept, their interpolations will highly maintain the semantic consistency of this concept (\ie, it ensures faithfulness). Meanwhile, since both images have different contexts, the interpolations will produce an image with a new context (\ie, it ensures diversity). Finally, we set a \emph{split ratio} to divide the whole denoising timesteps into two stages. In the first stage, we use a prompt containing the learned concept and a randomly sampled suffix\footref{footnote:appendix} to generate noisy images with diverse contexts (\eg, layout and gesture). In the second stage, we remove the suffix to refine the character details of the category concept. By adjusting the \emph{split ratio}, we can control the trade-off between faithfulness and diversity. To extract all suffixes, we first utilize a pretrained vision-language model to extract all captions of the original training set, then leverage a large language model to summarize them into a few suffixes.

We evaluated our method on various image classification tasks across multiple datasets and settings. Extensive results has demonstrated consistent improvements and significant gains over state-of-the-art methods. Conclusively, our contributions are summarized as follows:

\begin{itemize}[leftmargin=*]

\item We propose a unified view to analyze existing diffusion-based DA methods in scarce-data scenarios, we argue that they can not take account of both faithfulness and diversity well\footnote{More quantitative and qualitative analyses are left in the appendix.}, which results in limited improvements.

\item We propose an effective diffusion-based DA method, that leverages the inversion circle interpolation and two-stage denoising to generate faithful and diverse images.

\item Comprehensive evaluation on three tasks has verified that our Diff-II can achieve effective data augmentation by generating high-quality samples.
\end{itemize}
\section{Related Work}
\label{sec:2}

\noindent\textbf{Diffusion-Based DA.} 
With the emergence of diffusion models, diffusion-based DA~\citep{michaeli2024advancing,islam2024diffusemix} becomes a popular solution. One DA setting~\cite{yu2023diversify,azizi2023synthetic,he2022synthetic,zhou2023training} is enhancing \emph{coarse-grained} datasets. By tuning the diffusion model into the target domain~\cite{azizi2023synthetic} or leveraging the language model to generate general descriptions for characteristics~\cite{yu2023diversify}. The faithfulness of coarse-grained categories can be guaranteed. 

Meanwhile, another more crucial setting is enhancing small-scale \emph{fine-grained} datasets. Since it's more difficult to generate fine-grained appearances, DA methods always need to extract detailed patterns from reference images (\eg, an extra concept learning step). For this setting, there are two main paradigms: 1) Latent perturbation~\citep{zhou2023training, fu2024dreamda,zhang2024expanding} generate samples by perturbating latent codes in the latent space. Although these methods can generate diverse samples, due to the uncontrollable perturbation direction, the generated results sometimes deviate from the domain of the original dataset. Therefore, they heavily rely on extra over-sampling and filtering steps. 2) Image editing~\citep{he2022synthetic,trabucco2023effective,dunlap2023diversify,wang2024enhance} reformulate data augmentation as an image editing task~\citep{meng2021sdedit}. However, due to the limitations of the editing paradigm, it's difficult for them to take into account both the faithfulness and diversity of the synthetic samples. To tackle the above problem, our work proposes to generate new images by interpolating the inversions.

\noindent\textbf{Interpolation-Based DA.} For time series and text data, interpolation is a common approach for DA. Chen~\etal~\cite{chen2022doublemix} incorporate a two-stage interpolation in the hidden space to improve the text classification models. Oh~\etal~\cite{oh2020time} propose to augment time-series data by interpolation on original data. In the computer vision community, there are some studies~\citep{devries2017dataset,zhou2023training} work on interpolation-based DA for image classification. However, how to combine the excellent generation ability of diffusion models and interpolation operation to obtain high-quality synthetic samples remains an important challenge. In this paper, we utilize inversion circle interpolation by considering the distribution requirement for the diffusion model.

\section{Method}
\label{sec:3}

\textbf{Problem Formulation.}
For a general image classification task, typically there is a \textbf{original training set} with $K$ categories: $\mathcal{O} = \{\mathcal{O}^1, \mathcal{O}^2, ..., \mathcal{O}^K\}$, where $\mathcal{O}^i$ is the subset of all training samples belong to $i_{th}$ category. For $\mathcal{O}^i$, there are $N_i$ labeled training samples $\{X_j^i\}_{j=1}^{N_i}$. The classification task aims to train a classifier with $\mathcal{O}$ and evaluate it on the test set. On this basis, diffusion-based DA method first generates extra synthetic images for each category. The \textbf{Synthetic set}: $\mathcal{S} = \{\mathcal{S}^1, \mathcal{S}^2, ..., \mathcal{S}^K\}$, $\mathcal{S}^i$ is the set of synthetic images of $i_{th}$ category. Then it trains an improved classifier with both original and synthetic images (\ie, $\mathcal{O}\cup\mathcal{S}$).

\begin{figure*}[t]
    \centering
    \includegraphics[width=0.95\linewidth]{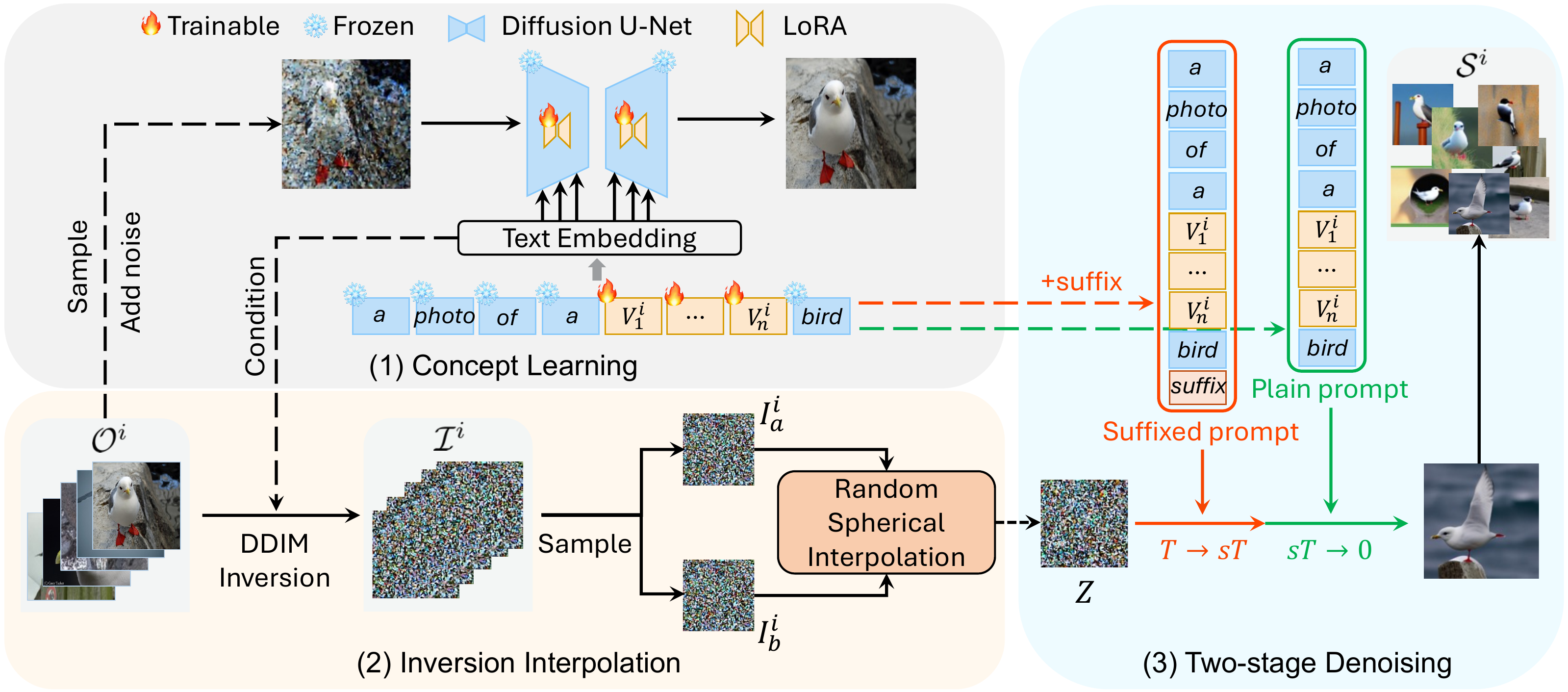}
    \vspace{-0.5em}
    \caption{\textbf{Pipeline of Diff-II}. (1) Concept Learning: Learning accurate concepts for each category. (2) Inversion Interpolation: Calculating DDIM inversion for each image conditioned on the learned concept. Then, randomly sampling a pair and conducting random circle interpolation. (3) Two-stage Denoising: Denosing the interpolation results in a two-stage manner with different prompts.}
    \label{fig:3}
\end{figure*}

\textbf{General Framework.} As shown in Figure~\ref{fig:3}, our proposed Diff-II consists of three main steps: 

\noindent 1) \underline{Category Concepts Learning}: We first set $n$ learnable token embeddings for each category, and insert some learnable low-rank matrixes into the pretrained diffusion U-Net. By reconstructing the noised image of the original training set $\mathcal{O}$, we learn the accurate concept for each category. We denote the tokens of the $i_{th}$ category concept as $\{[V_{j}^i]\}_{j=1}^{n}$.

\noindent 2) \underline{Inversion Interpolation}: Take the $i_{th}$ category as an example, we form a prompt: ``a photo of a $[V_{1}^i]$ $[V_{2}^i]$ ... $[V_{n}^i]$ \texttt{[metaclass]}\textsuperscript{\ref{footnote:appendix}}". The ``\texttt{[metaclass]}" is the theme of the corresponding dataset, \eg ``bird" is the ``\texttt{[metaclass]}" for dataset CUB~\citep{wah2011caltech}. Then, we calculate the DDIM inversion $I_j^i$ for each training sample $X_j^i\in\mathcal{O}^i$ conditioned on this prompt. All these inversions (from $\mathcal{O}^i$) made up the \textbf{inversion pool} $\mathcal{I}^i=\{I_j^i\}_{j=1}^{N_i}$ (\cf, Sec.~\ref{sec:3.2.1}).
After that, we randomly sample two inversions ($I_a^i$, $I_b^i$) from $\mathcal{I}^i$ and conduct \textbf{random circle interpolation} on this pair (\cf, Sec.~\ref{sec:3.2.2}). The interpolation result is denoted as $Z$. We repeat the sampling and interpolation then collect all interpolation results into $\mathcal{Z}^i$.

\noindent 3) \underline{Two-stage Denoising}: Given an interpolation $Z\in\mathcal{Z}^i$, we denoise it as the initial noise in two stages. The main difference between the two stages is the prompt used. In the first stage, we use a \textbf{suffixed prompt}: ``a photo of a $[V_{1}^i]$ $[V_{2}^i]$ ... $[V_{n}^i]$ \texttt{[metaclass]} \texttt{[suffix]}". In the second stage, we use a \textbf{plain prompt}: ``a photo of a $[V_{1}^i]$ $[V_{2}^i]$ ... $[V_{n}^i]$ \texttt{[metaclass]}". Repeat two-stage denoising for each $Z\in\mathcal{Z}^i$, then we can get all the synthetic images and collect them into $\mathcal{S}^i$.

\subsection{Category Concepts Learning}
\label{sec:3.1}

The pre-trained datasets of diffusion models may have a distribution gap with the downstream classification benchmarks. Thus, it is hard to directly use category labels to guide the diffusion model to generate corresponding faithful images. Learning a more faithful concept
 for each category as the prompt for downstream generation is quite necessary. To achieve this, we followed the same learning strategy as~\citep{wang2024enhance}. Specifically, there are two learnable parts: 1) \emph{Token embeddings}: For the $i_{th}$ category, we set $n$ learnable concept tokens ($\{[V_{j}^i]\}_{j=1}^{n}$) 2) \emph{Low-rank matrixes}: We insert some low-rank matrixes~\citep{hu2021lora} into the pretrained diffusion U-Net. These matrixes are shared by all categories.

Based on the above, given $X_j^i\in\mathcal{O}^i$, its prompt is ``a photo of a $[V_{1}^i]$ $[V_{2}^i]$ ... $[V_{n}^i] $\texttt{[metaclass]}". For timestep $t$ in the forward process of diffusion, the noised latent $x_t$ can be calculated as follows:
\begin{equation}
    x_t=\sqrt{\Bar{\alpha}_t}x_0+\sqrt{1-\Bar{\alpha}_t}\epsilon,
    \label{eq:1}
\end{equation}
where $x_0$ is the encoded latent of $X_j^i$, $\Bar{\alpha}_t$ is a pre-defined parameter and $\epsilon$ is a Gaussian noise. Training objective is:
\begin{equation}
    \mathop{\mathrm{min}}_{\theta} \mathbb{E}_{\epsilon,x,c,t} \left[||\epsilon-\epsilon_\theta(x_t,c,t)||_2^2 \right],
    \label{eq:2}
\end{equation}
where $c$ is the encoded prompt, $\epsilon_\theta$ is the predicted noise of the diffusion model.

\subsection{Inversion Interpolation}
\label{sec:3.2}

\subsubsection{Inversion Pool Construction}
\label{sec:3.2.1}

To get a faithful and diverse synthetic set by interpolating image pairs, we propose to conduct interpolation in the DDIM~\citep{song2020denoising} inversion space\footnote{More backgrounds about DDIM and DDIM inversion are in appendix}. There are two main motivations: 1) The sampling speed of DDIM is competitive due to the sampling of non-consecutive time steps. This can make our inverse process efficient. 2) We found that starting from the DDIM inversion can ensure a relatively high reconstruction result, especially conditioned on the learned concepts from Sec.~\ref{sec:3.1}. 

DDIM sampling has the following updating equation:
\begin{small} 
\begin{equation}
\begin{split}
    x_{t-1}=\sqrt{\Bar{\alpha}_{t-1}}(\frac{x_t-\sqrt{1-\Bar{\alpha}_t}\epsilon_\theta(x_t,c,t)}{\sqrt{\Bar{\alpha}}_{t}})\\+ \sqrt{1-\Bar{\alpha}_{t-1}}\epsilon_\theta(x_t,c,t),    
\end{split}
    \label{eq:3}
\end{equation}
\end{small}
where $x_t$ is the latent at timestep $t$ in reverse process.
Based on Eq.~(\ref{eq:3}) and $\theta(x_{t},c,t) \simeq \theta(x_{t-1},c,t)$ , we can get the DDIM inversion update equation:

\begin{small} 
\begin{equation}
\begin{split}
    x_{t}\simeq\frac{\sqrt{\Bar{\alpha}_{t}}}{(\sqrt{\Bar{\alpha}_{t}}-1)}(x_{t-1}-\sqrt{1-\Bar{\alpha}_{t-1}}\epsilon_\theta(x_{t-1},c,t))\\ +
    \sqrt{1-\Bar{\alpha}_{t}}\epsilon_\theta(x_{t-1},c,t)
    \label{eq:4}
\end{split}
\end{equation}
\end{small}
Given a training sample $X^i_j\in\mathcal{O}^i$, we first encode it into $x_0$ with a VAE encoder. Then we leverage Eq.~(\ref{eq:4}) to update $x_t$ while $c$ is the text embedding of ``a photo of a $[V_{1}^i]$ $[V_{2}^i]$ ... $[V_{n}^i]$ \texttt{[metaclass]}". When $t$ reaches the maximum timestep $T$, the $x_T$ is the final DDIM inversion. After conducting Eq.~(\ref{eq:4}) for each $X^i_j\in\mathcal{O}^i$, we can construct an \textbf{inversion pool} $\mathcal{I}^i=\{I^i_j\}_{j=1}^{N_i}$. 

\subsubsection{Random Circle Interpolation}
\label{sec:3.2.2}

Since Gaussian noises are received as input during the training process of the diffusion model, we need to ensure the initial noise for the denoising process also resides in a Gaussian distribution. Since each inversion in $\mathcal{I}^i$ is in a Gaussian distribution, the common linear interpolation will lead to a result that is not in Gaussian distribution.
Thus, we propose to conduct circle interpolation on the inversion pairs. This operation has a larger interpolation range (which increases the diversity) and can maintain the interpolation result in Gaussian distribution\textsuperscript{\ref{footnote:appendix}}. Thus, it can be the initial noise for the denoising process.

After getting the inversion set $\mathcal{I}^i$, we randomly select two DDIM inversions $I_a, I_b$ (ignored the superscript) as a pair from $\mathcal{I}^i$. For this pair, we conduct the random circle interpolation.

\textbf{Circle interpolation.} The circle interpolation can be intuitively understood as rotating from one to another and it can be expressed as follows:
\begin{small} 
\begin{equation}
        Z = \frac{sin((1+\lambda)\alpha)}{sin(\alpha)}I_a-\frac{sin(\lambda\alpha)}{sin(\alpha)}I_b, \qquad 
 \lambda\in[0,\frac{2\pi}{\alpha}],
    \label{eq:7}
\end{equation}
\end{small}
where $\alpha=arccos(\frac{I_a^TI_b}{(||I_a||||I_b||))})$ and $Z$ is the interpolation result. $\lambda$ is a random interpolation strength, which can decide the interpolation type (interpolation or extrapolation) and control the relative distance between $Z$ and $I_a$, $I_b$. 

As shown in Figure~\ref{fig:4}, the path of circle interpolation is the circle composed of the \textcolor{ForestGreen}{Green Arc} and the \textcolor{blue}{Blue Arc}.
According to the rotation direction, we can decompose the circle interpolation into spherical interpolation and spherical extrapolation~\citep{shoemake1985animating}:

\begin{figure}[t]
    \centering
    \includegraphics[width=0.8\linewidth]{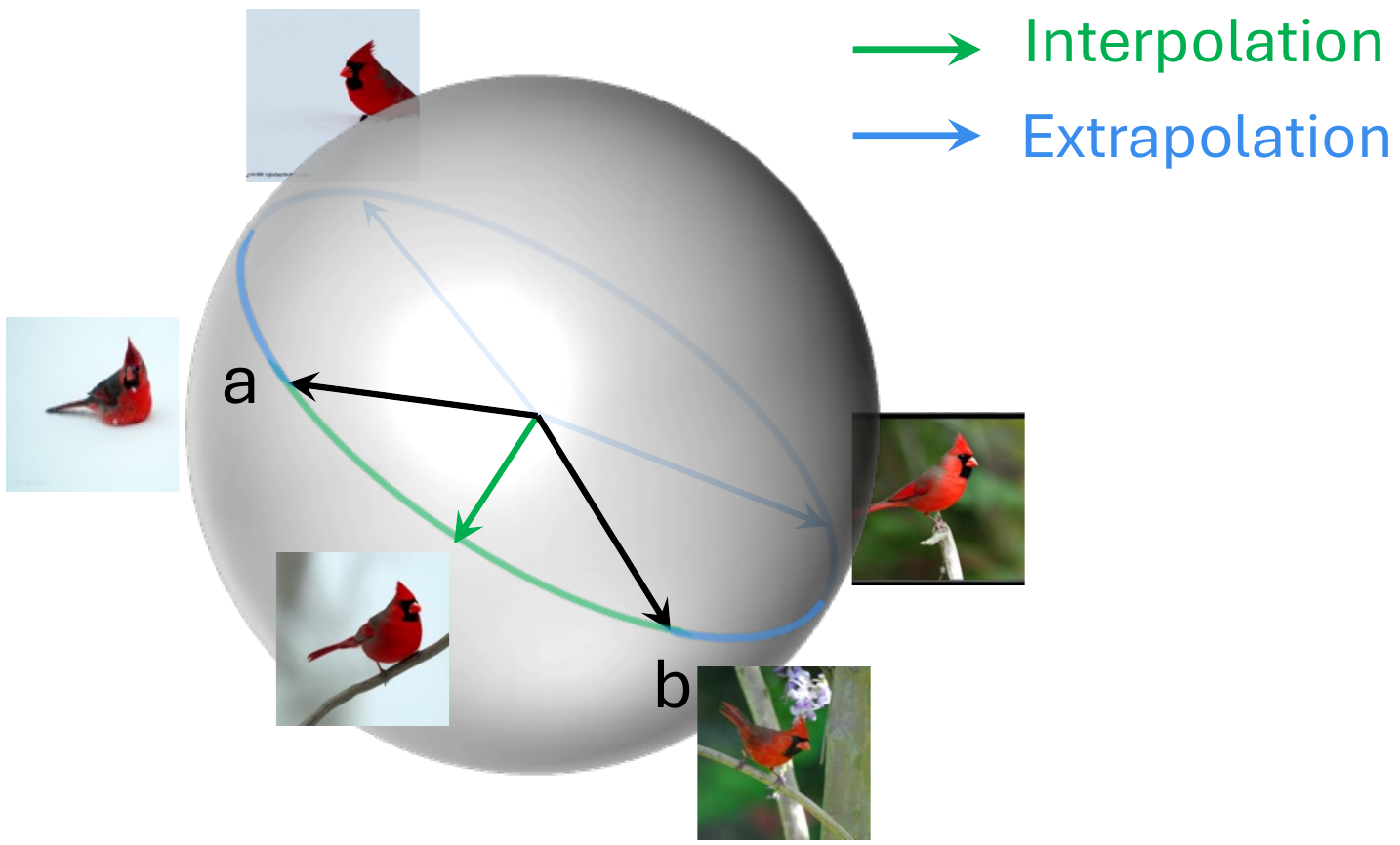}
    \vspace{-0.5em}
    \caption{An illustration for the proposed circle interpolation.}
    \label{fig:4}
\end{figure}

\textbf{Spherical Interpolation.} The spherical interpolation means rotate along the shortest path (\cf, the \textcolor{ForestGreen}{Green Arc} of Figure~\ref{fig:4}) and it can be expressed as follows:
\begin{small} 
\begin{equation}
    Z = \frac{sin((1-\lambda)\alpha)}{sin(\alpha)}I_a+\frac{sin(\lambda\alpha)}{sin(\alpha)}I_b    ,\qquad     
      \lambda\in[0,1]
    \label{eq:5}
\end{equation}
\end{small}


\textbf{Spherical Extrapolation.} The spherical extrapolation means rotate along the opposite direction of the interpolation path (\cf, the \textcolor{blue}{Blue Arc} of Figure~\ref{fig:4}) and it can be expressed as follows:
\begin{small} 
\begin{equation}
    Z = \frac{sin((1+\lambda)\alpha)}{sin(\alpha)}I_a-\frac{sin(\lambda\alpha)}{sin(\alpha)}I_b,\qquad     
      \lambda\in[0,\frac{2\pi}{\alpha}-1]
    \label{eq:6}
\end{equation}
\end{small}
According to the periodicity of trigonometric functions, we can see that Eq.~(\ref{eq:7}) is a unified representation of spherical interpolation (Eq.~(\ref{eq:5})) and spherical extrapolation (Eq.~(\ref{eq:6})). Based on the expansion rate of the $i_{th}$ category, we repeat the sampling and interpolation. Then, we collect all the interpolation results into $\mathcal{Z}^i$, which will be used as the initial noises in Sec.~\ref{sec:3.3}.

\subsection{Two-stage Denoising}
\label{sec:3.3}
\begin{figure*}[t]
    \centering
    \includegraphics[width=0.95\linewidth]{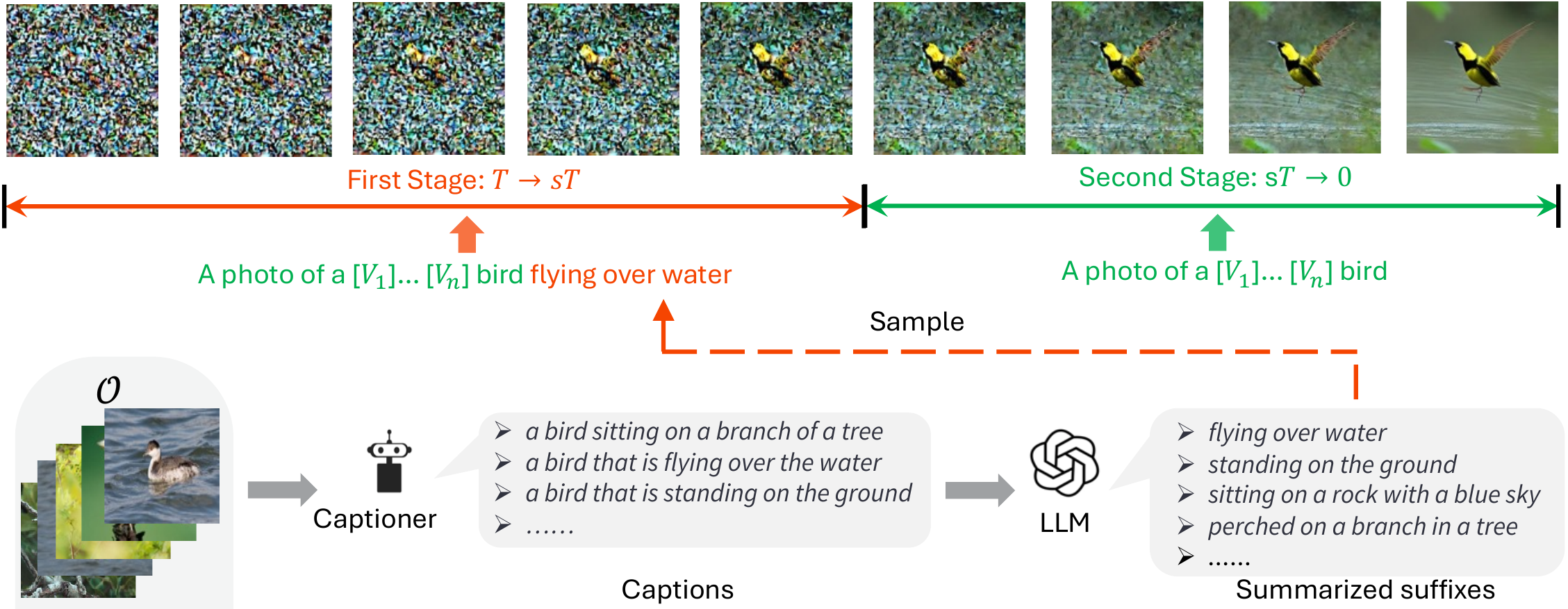}
    \vspace{-0.5em}
    \caption{\textbf{Two-stage denoising}. Input all images into a captioner and get all captions. Then leverage the language model to summarize these captions into some suffixes. Finally, denoise with the suffixed prompt in the first stage and with the plain prompt in the second stage.}
    \label{fig:5}
    \vspace{-0.4em}
\end{figure*}

In order to further increase the diversity of synthetic images, we design a two-stage denoising process (\cf, Figure~\ref{fig:5}). We split the denoising process into two stages with a \textbf{split ratio} $s\in[0,1]$. The first stage includes time steps from $T$ to $sT$. The second stage includes time steps from $sT$ to $0$. The main difference between the two stages is the prompt used.

\textbf{Suffixed Prompt.} For a specific dataset, we will generate a few suffixes that can summarize the context of this dataset. First, we input each $X_j\in\mathcal{O}$ into a pre-trained vision language model (VLM) (\eg, BLIP~\citep{li2022blip}) to get the corresponding caption. After getting all the captions, we input them into a large language model (LLM) (\eg, GPT-4~\citep{achiam2023gpt}) to summarize them into a few descriptions with the following format: ``a photo of a \texttt{[metaclass]} \texttt{[suffix]}". Thus, we can get a few suffixes for a dataset\textsuperscript{\ref{footnote:appendix}}. Based on the captioning ability of VLM and the powerful generalization ability of LLM, these suffixes summarize the high-frequency context in the dataset. For each $Z\in\mathcal{Z}^i$, we randomly sample one suffix then concat the plain prompt with this suffix into: ``a photo of a $[V_{1}^i]$ $[V_{2}^i]$ ... $[V_{n}^i]$ \texttt{[metaclass]} \texttt{[suffix]}".

\textbf{Denoising Process.} Based on the above, the first stage uses the suffixed prompt while the second stage removes the suffix part. We can express our two-stage denoising process as follows:
\begin{equation}
\begin{split}
    x_{t-1}=\sqrt{\Bar{\alpha_{t-1}}}(\frac{x_t-\sqrt{1-\Bar{\alpha_t}}\epsilon_\theta}{\sqrt{\Bar{\alpha_{t}}}})+\sqrt{1-\Bar{\alpha_{t-1}}}\epsilon_\theta \quad 
\end{split}
    \label{eq:8} 
\end{equation}
where $x_T=Z\in\mathcal{Z}^i$, $c^*$ and $c$ are the text embedding of suffixed prompt and the prompt without suffix part respectively. After the above update, we can obtain $x_0$ and then form the synthetic set $\mathcal{S}$.




\section{Experiments}
\label{sec:4}

\subsection{Few-shot Classification}
\label{sec:4.1}

\begin{table*}[t]
\resizebox{2\columnwidth}{!}{%
\def\arraystretch{0.95}
\begin{tabular}{clccccccccccc}
\hline
\hline
\multicolumn{2}{c}{\multirow{2}{*}{\large{Method}}} &
  \multicolumn{5}{c}{\textbf{5-shot}} &
   &
  \multicolumn{5}{c}{\textbf{10-shot}} \\ \cline{3-7} \cline{9-13} 
\multicolumn{2}{c}{} &
  CUB &
  Aircraft &
  Pet &
  Car &
  Avg &
   &
  CUB &
  Aircraft &
  Pet &
  Car &
  Avg \\ \hline  
\multirow{8}{*}{\rotatebox{90}{ResNet50}} &
  Original & 
  \multicolumn{1}{c}{54.52} &
  \multicolumn{1}{c}{36.63} &86.98
   & 40.10
   & 54.56
   &
   & 70.26
   & 54.11
   & 90.41
   & 69.64
   & 71.11
   \\
 &
   Mixup\citep{zhang2017mixup}  &
  \multicolumn{1}{c}{54.33} &
  \multicolumn{1}{c}{36.75} & 86.36
   & 41.43
   & 54.72\color{ForestGreen}\small{$(+0.16)$}
   &
   & 70.00
   & 55.28
   & 89.49
   & 70.28
   & 71.26\color{ForestGreen}\small{$(+0.15)$}
   \\
 &
   CutMix\citep{yun2019cutmix}  &
  \multicolumn{1}{c}{51.74} & 
  \multicolumn{1}{c}{33.09} & 86.52
   & 36.19
   & 51.89\color{red}\small{$(-2.67)$}
   &
   & 68.91
   & 51.05
   & 89.71
   & 67.71
   & 69.35\color{red}\small{$(-1.76)$}
   \\ 
 &
  Real-Filter~\citep{he2022synthetic} &
  \multicolumn{1}{c}{55.03} &
  \multicolumn{1}{c}{30.40} & 87.66
   & 57.50
   & 57.65\color{ForestGreen}\small{$(+3.09)$}
   &
   & 67.36
   & 47.21
   & 90.14
   & 69.64
   & 70.66\color{red}\small{$(-0.45)$}
   \\
 &
  Real-Guidance~\citep{he2022synthetic} &
  \multicolumn{1}{c}{54.67} &
  \multicolumn{1}{c}{36.27} & 86.22
   & 42.45
   & 54.90\color{ForestGreen}\small{$(+0.34)$}
   &
   & 69.81
   & 54.77
   & 90.06
   & 70.97
   & 71.40\color{ForestGreen}\small{$(+0.29)$}
   \\
 &
  Real-Mix~\citep{wang2024enhance} &
  \multicolumn{1}{c}{51.00} &
  \multicolumn{1}{c}{22.53} & 88.63
   & 52.06
   & 53.56\color{red}\small{$(-1.00)$}
   & 
   & 64.65
   & 44.16
   & 89.40
   & 75.72
   & 68.48\color{red}\small{$(-2.63)$}
   \\
 &
  Da-Fusion~\citep{trabucco2023effective} & 59.40
   &   \multicolumn{1}{c}{34.98}
   & 88.64
   & 51.90
   & 58.73\color{ForestGreen}\small{$(+4.17)$}
   & 
   & 72.05
   & 51.14
   & 90.47
   & 77.61
   & 72.82\color{ForestGreen}\small{$(+1.71)$}
   \\
 &
  Diff-AUG~\citep{wang2024enhance} & 61.14
   & \multicolumn{1}{c}{39.48}
   & 89.24
   & 62.28
   & 63.04\color{ForestGreen}\small{$(+8.78)$}
   & 
   & 72.02
   & 55.43
   & 90.28
   & 81.65
   & 74.85\color{ForestGreen}\small{$(+3.74)$}
   \\
 &
  Diff-Mix~\citep{wang2024enhance} & 
  \multicolumn{1}{c}{56.18} &
  \multicolumn{1}{c}{32.61} & 88.77
   & 56.39
   & 58.49\color{ForestGreen}\small{$(+3.93)$}
   &
   & 70.16
   & 52.33
   & 90.79
   & 78.68
   & 72.99\color{ForestGreen}\small{$(+1.88)$}
   \\
 &
  \textbf{Ours} &
   \cellcolor{mygray}\textbf{62.22}&
   \multicolumn{1}{c}{\cellcolor{mygray}\textbf{42.15}}&
   \cellcolor{mygray}\textbf{89.84}&
   \cellcolor{mygray}\textbf{64.24}&
  \cellcolor{mygray}\textbf{64.61}\color{ForestGreen}\small{$(+10.05)$} &
  \cellcolor{mygray}\textbf{} &
  \cellcolor{mygray}\textbf{72.66} &
  \cellcolor{mygray}\textbf{57.43} &
  \cellcolor{mygray}\textbf{91.01} &
  \cellcolor{mygray}\textbf{82.02} &
  \cellcolor{mygray}\textbf{75.78}\color{ForestGreen}\small{$(+4.67)$} \\ \hline
\multirow{8}{*}{\rotatebox{90}{ViT-B/16}} &
  Original & 73.00
   & \multicolumn{1}{c}{32.85}
   & 90.71
   & 59.72
   & 64.07
   &
   & 83.52
   & 51.59
   & 92.92
   & 81.01
   & 77.26
   \\
 &Mixup\citep{zhang2017mixup}
  &
  \multicolumn{1}{c}{76.07} &
  \multicolumn{1}{c}{36.15} & 90.77
   & 61.92
   & 66.23\color{ForestGreen}\small{$(+2.16)$}
   &
   & 85.24
   & 52.94
   & 92.81
   & 82.46
   & 78.36\color{ForestGreen}\small{$(+1.10)$}
   \\
 &
   CutMix\citep{yun2019cutmix}  &
  \multicolumn{1}{c}{72.97} &
  \multicolumn{1}{c}{32.79} & 90.22
   & 59.16
   & 63.79\color{red}\small{$(-0.28)$}
   &
   & 83.95
   & 51.56
   & 92.59
   & 81.71
   & 77.45\color{ForestGreen}\small{$(+0.19)$}
   \\
 &
  Real-Filter~\citep{he2022synthetic} & 73.66
   & \multicolumn{1}{c}{31.44}
   & 90.85
   & 72.51
   & 67.12\color{ForestGreen}\small{$(+3.05)$}
   & 
   & 82.19
   & 46.43
   & 93.14
   & 84.69
   & 76.61\color{red}\small{$(-0.65)$}
   \\
 &
  Real-Guidance~\citep{he2022synthetic} & 74.74
   & \multicolumn{1}{c}{35.10}
   & 91.23
   & 62.34
   & 65.85\color{ForestGreen}\small{$(+1.78)$}
   & 
   & 83.24
   & 53.15
   & 93.22
   & 81.97
   & 77.90\color{ForestGreen}\small{$(+0.64)$}
   \\
 &
  Real-Mix~\citep{wang2024enhance} & 72.25
   & \multicolumn{1}{c}{32.91}
   & 90.92
   & 70.55
   & 66.66\color{ForestGreen}\small{$(+2.59)$}
   &
   & 80.95
   & 47.01
   & 93.16
   & 83.78
   & 76.23\color{red}\small{$(-1.03)$}
   \\
 &
  Da-Fusion~\citep{trabucco2023effective} & 76.24
   & \multicolumn{1}{c}{34.20}
   & 93.03
   & 71.07
   & 68.64\color{ForestGreen}\small{$(+4.57)$}
   &
   & 83.97
   & 51.68
   & 93.68
   & 85.00
   & 78.58\color{ForestGreen}\small{$(+1.32)$}
   \\
 &
  Diff-AUG~\citep{wang2024enhance} & 77.24
   & \multicolumn{1}{c}{40.08}
   & 92.27
   & 76.48
   & 71.52\color{ForestGreen}\small{$(+7.45)$}
   &
   & 84.10
   & 56.24
   & 93.93
   & 87.59
   & 80.47\color{ForestGreen}\small{$(+3.21)$}
   \\
 &
  Diff-Mix~\citep{wang2024enhance} &
  \multicolumn{1}{c}{74.49} &
  \multicolumn{1}{c}{36.48} & 92.29
   & 73.01
   & 69.07\color{ForestGreen}\small{$(+5.00)$}
   &
   & 81.98
   & 53.92
   & 93.98
   & 85.42
   & 78.83\color{ForestGreen}\small{$(+1.57)$}
   \\
 &
  \textbf{Ours} &
   \cellcolor{mygray}\textbf{79.22}&
   \multicolumn{1}{c}{\cellcolor{mygray}\textbf{40.23}}&
   \cellcolor{mygray}\textbf{93.79}&
   \cellcolor{mygray}\textbf{77.10}&
   \cellcolor{mygray}\textbf{72.59}\color{ForestGreen}\small{$(+8.52)$}&
   \cellcolor{mygray}\textbf{}&
   \cellcolor{mygray}\textbf{84.89}&
   \cellcolor{mygray}\textbf{56.32}&
   \cellcolor{mygray}\textbf{94.25}&
   \cellcolor{mygray}\textbf{87.81}&\cellcolor{mygray}\textbf{80.82}\color{ForestGreen}\small{$(+3.56)$}
   \\ \hline\hline
\end{tabular}%
}
\vspace{-1em}
\caption{\textbf{Few-shot classification}. 5-shot and 10-shot results (averaged on three trials) on four fine-grained datasets with two backbones. ``Original" means the model trained on the original set w/o DA. \textcolor{ForestGreen}{Green} and \textcolor{red}{red} numbers are increase and decrease values w.r.t. ``Original".}
\label{table:1}
\vspace{-0.2cm}
\end{table*}

\noindent\textbf{Settings.}
To evaluate the Diff-II's augmentation capacity based on few samples, we conducted few-shot classification on four domain-specific fine-grained datasets: \emph{CUB}~\citep{wah2011caltech}, \emph{Aircraft}~\citep{maji2013fine}, \emph{Cars}~\citep{krause20133d} and \emph{Pet}~\citep{parkhi2012cats}, with shot numbers of 5, 10. We used the augmented datasets to fine-tune two backbones: 224$\times$224-resolution ResNet-50~\citep{he2016deep} pre-trained on ImageNet1K~\citep{deng2009imagenet} and 384$\times$384 ViT-B/16~\citep{dosovitskiy2020image} pre-trained on ImageNet21K. We compared our method with two DA augmentation methods: \emph{Mixup}~\cite{zhang2017mixup}, \emph{CutMix}~\citep{yun2019cutmix} and six diffusion-based DA methods: \emph{Real-Filter}, \emph{Real-Guidance}~\citep{he2022synthetic}, \emph{Da-Fusion}~\citep{trabucco2023effective}, \emph{Real-Mix}, \emph{Diff-AUG} and \emph{Diff-Mix}~\citep{wang2024enhance}. We fixed $s$ to 0.3 for 5-shot and 0.1 for 10-shot.
For fairness, the expansion rate was 5 for all methods. For the classifier training process, we followed the joint training strategy of ~\citep{trabucco2023effective}: replacing the data from the original set with synthetic data in a replacement probability during training. We fixed the replacement probability with 0.5 for all methods. More details are in the \textbf{Appendix}.

\noindent\textbf{Results.} 
From the results in Table~\ref{table:1}, we have several observations: 1) Compared with training on the original set, our method can improve the average accuracy from $3.56\%$ to $10.05\%$, indicating that our methods can effectively augment domain-specific fine-grained datasets. 2) Our method can outperform all the comparison methods in all settings, demonstrating the effectiveness of our method for few-shot scenarios. 3) Our method achieves greater gains for smaller shots (\ie, 5-shot) and weaker backbone (\ie, ResNet-50), showing our method is robust to challenging settings.

\subsection{Long-tail Classification}
\label{sec:4.2}

\noindent\textbf{Settings.} 
To evaluate the Diff-II's augmentation capacity for datasets with imbalanced samples, we experimented with our methods on the long-tail classification task. Following the previous settings~\citep{cao2019learning,liu2019large,park2022majority,wang2024enhance}, we evaluated our method on two domain-specific long-tail datasets: \emph{CUB-LT}~\citep{samuel2021generalized} and \emph{Flower-LT}~\citep{wang2024enhance}, with imbalance factor (IF) of 100, 20, and 10. 
We used the 224$\times$224-resolution ResNet-50 (mentioned in Sec.~\ref{sec:4.1}) as the backbone. We fixed $s$ to 1.0 for all settings. For those categories with only one image, we randomly sample a noise for subsequent generation since we can not interpolate.
We compared our method with five methods: \emph{oversampling-based CMO}~\citep{park2022majority}, \emph{re-weighting CMO}~\citep{cao2019learning}, diffusion-based \emph{Real-Filter}, \emph{Real-Guidance}~\cite{he2022synthetic}, \emph{Real-Mix}, and \emph{Diff-Mix}~\cite{wang2024enhance}. For fairness, the expansion rate was 5, and the replacement probability was 0.5 for all diffusion-based methods. More details are in the \textbf{Appendix}.

\addtolength{\tabcolsep}{-4pt}
\begin{table}[t]
\resizebox{\columnwidth}{!}{%
\def\arraystretch{1.1}
\vspace{-1em}
\begin{tabular}{lccccccccc}
\hline\hline
\multirow{2}{*}{Method} & \multicolumn{4}{c}{CUB-LT}     &  & \multicolumn{4}{c}{Flower-LT}  \\ \cline{2-5} \cline{7-10} 
                        & IF=100 & IF=20 & IF=10 & Avg   &  & IF=100 & IF=20 & IF=10 & Avg   \\ \hline
CE                      & 33.65  & 44.82 & 58.13 & 45.53 &  & 80.43  & 90.87 & 95.07 & 88.79 \\
CMO~\citep{park2022majority}                     & 32.94  & 44.08 & 57.62 & \color{red}{$-0.65$} &  & 83.95  & 91.43 & 95.19 & \color{ForestGreen}{$+1.40$} \\
CMO-DRW~\citep{cao2019learning}                 & 32.57  & 46.43 & 59.25 & \color{ForestGreen}{$+0.55$} &  & 84.07  & 92.06 & 95.92 & \color{ForestGreen}{$+1.89$} \\
Real-Gen~\citep{wang2024enhance}                & 45.86  & 53.43 & 61.42 & \color{ForestGreen}{$+8.04$} &  & 83.56  & 91.84 & 95.22 & \color{ForestGreen}{$+1.42$} \\
Real-Mix~\citep{wang2024enhance}                & 47.75  & 55.67 & 62.27 & \color{ForestGreen}{$+9.70$} &  & 85.19  & 92.96 & 96.04 & \color{ForestGreen}{$+2.61$} \\
Diff-Mix~\citep{wang2024enhance}                & 50.35  & 58.19 & 64.48 & \color{ForestGreen}{$+12.14$} &  & 89.46  & 93.86 & 96.63 & \color{ForestGreen}{$+4.53$} \\
\textbf{Ours} & \cellcolor{mygray}\textbf{51.21} & \cellcolor{mygray}\textbf{62.31} & \cellcolor{mygray}\textbf{70.28} & \cellcolor{mygray}\color{ForestGreen}{$+15.74$} &  & \cellcolor{mygray}\textbf{89.54} & \cellcolor{mygray}\textbf{94.39} & \cellcolor{mygray}\textbf{97.35} & \cellcolor{mygray}\color{ForestGreen}{$+4.97$} \\ \hline\hline
\end{tabular}%
}
\vspace{-1em}
\caption{\textbf{Long-tail classification results on CUB-LT and Flower-LT}. ``CE'' is a plain baseline that trains a classifier on the original set with the Cross-Entropy loss. It contains no operations designed for long-tail tasks. ``IF'' is the imbalanced factor, where a larger IF indicates more imbalanced data distribution. \textcolor{ForestGreen}{Green} and \textcolor{red}{red} numbers are the increase and decrease values w.r.t. CE. ``Ours" results are averaged on three trials, and other results are from~\citep{wang2024enhance}} 
\vspace{-1em}
\label{table:2}
\end{table}
\addtolength{\tabcolsep}{4pt}

\noindent\textbf{Results.} 
From the results in Table~\ref{table:2}, we have several observations: 1) Our method can outperform all the comparison methods in all settings. For example, the average accuracy on \emph{CUB-LT} exceeds the previous state-of-the-art \emph{Diff-Mix} $3.6\%$,  demonstrating our method can well mitigate the imbalanced data distribution. 2) Compared with the case of relatively low imbalanced factors (\eg, IF=10), the gain brought by our method will be reduced when the imbalanced factor is quite high (\eg, IF=100). This is because when the imbalance is too high, there is only one sample for many categories, making our inversion interpolation can not be implemented.

\subsection{Out-of-Distribution (OOD) Classification}
\label{sec:4.3}

\noindent\textbf{Settings.} 
To evaluate whether the synthetic data generated by Diff-II can benefit the generalization capacity of the classifier, we conducted OOD classification experiments. To be specific, we trained a 224$\times$224-resolution ResNet-50 (\cf, Sec.~\ref{sec:4.1}) with the original set of \emph{5-shot CUB} and corresponding synthetic data (same with Sec.~\ref{sec:4.1}, and evaluated it on a prevalent OOD dataset: \emph{Waterbird}~\citep{sagawa2019distributionally}. 
Besides, the comparison methods were six diffusion-based data augmentation methods: \emph{Real-Filter}~\cite{he2022synthetic}, \emph{Real-Guidance}~\cite{he2022synthetic}, \emph{Da-Fusion}~\cite{trabucco2023effective}, \emph{Real-Mix}~\cite{wang2024enhance}, \emph{Diff-AUG}~\cite{wang2024enhance} and \emph{Diff-Mix}~\cite{wang2024enhance}. We used the same hyper-parameters with Sec.~\ref{sec:4.1}.

\noindent\textbf{Results.} 
As shown in Table~\ref{table:3}, we can have two observations: 1) Our method can significantly improve the classification ability of the classifier on the background-shift out-of-distribution dataset by augmenting the original dataset. For example, the average accuracy can be improved by $11.39\%$ compared to the ``Original'' (no augmentation) one. This shows that the data generated by our Diff-II has good diversity, so it is possible to train a classifier that is robust to the background. 2) Our method can outperform all the comparison methods in 4 groups. Especially in (water, land) group, Diff-II can outperform the second-best method (\emph{Diff-AUG}) by $3.45\%$. This demonstrates the excellent ability of our method to generate faithful and diverse images.

\subsection{Ablation Study}
\label{sec:4.4}

\begin{figure*}[t]
    \centering
    \includegraphics[width=1\linewidth]{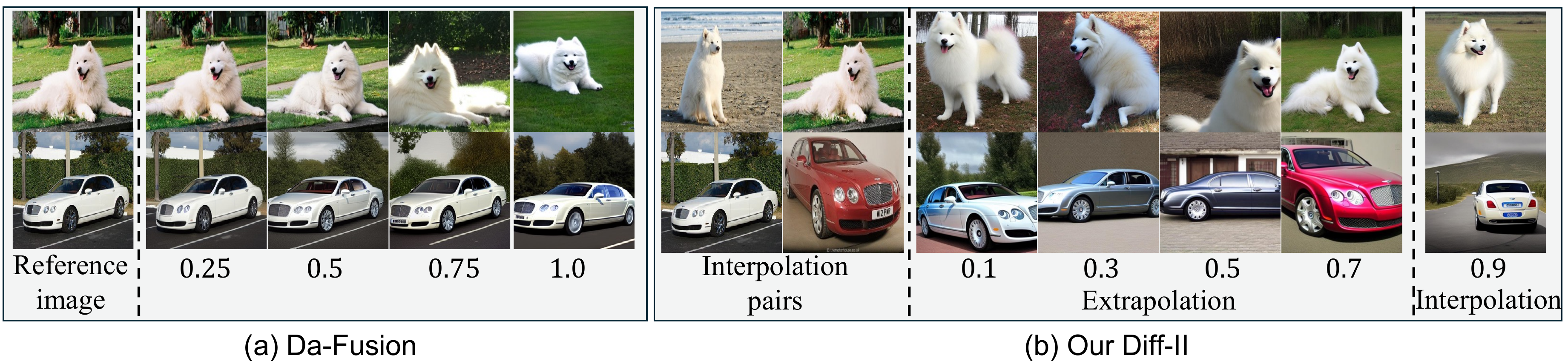}
    \vspace{-2em}
    \caption{\textbf{Visualization Comparison}. (a) Synthetic images of Da-fusion regarding different translation strengths. (b) Synthetic images of our Diff-II regarding different interpolation strengths (The unit is $2\pi/\alpha$). Experientially, the interpolation type is extrapolation when the strength is in $[0,0.75]$, else interpolation.}
    \label{fig:7}
    \vspace{-1em}
\end{figure*}

\textbf{Effectiveness of Each Component.}
We investigated the effectiveness of each component on the \emph{5-shot Aircraft} (same setting as Sec.~\ref{sec:3.1} with ResNet) and reported: average \emph{LPIPS}~\citep{zhang2018unreasonable} between images of the synthetic set (which can reflect the diversity), and classification \emph{accuracy}. As shown in Table~\ref{table:4}, the first row indicates denoising a random noise with a templated prompt (\eg, ``a photo of a Samoyed pet''). Since it lacks \texttt{CL}, the accuracy is quite poor, which indicates that for fine-grained datasets, concept learning is a crucial component. 
By comparing other rows, we can conclude that each component of our method (\ie, \texttt{CL}, \texttt{SI}, \texttt{SE}, \texttt{TD}) can improve the diversity and final classification performance. After combining all components, the \emph{LPIPS} further increased, thus boosting higher \emph{accuracy}.

\addtolength{\tabcolsep}{-2pt}
\begin{figure}[t]
\vspace{-0.5em}
\resizebox{1.0\columnwidth}{!}{%
\def\arraystretch{1}
\begin{tabular}{lccccc}
\hline
\hline
Method        & \multicolumn{1}{c}{L,L} & \multicolumn{1}{c}{W,W} & \multicolumn{1}{c}{L,W} & \multicolumn{1}{c}{W,L} & Avg \\ \hline
Original      &  38.55                    &    32.15                  &   37.90        &   30.20        & 34.70 \\
Real-Filter~\cite{he2022synthetic}   &  38.87                    &  34.72                    &   38.25        &    31.64       & 35.87\color{ForestGreen}\small{$(+1.17)$} \\
Real-Guidance~\cite{he2022synthetic} &   39.49                   & 33.64                     & 39.80          &   29.78        &  35.68\color{ForestGreen}\small{$(+0.98)$}\\
Real-Mix~\cite{wang2024enhance}      & 30.29                     &  29.44                    &   30.02        &   26.95        & 29.18\color{red}\small{$(-5.52)$} \\
Da-Fusion~\cite{trabucco2023effective}     & 44.50 & 39.04 &     44.46      &  33.49         & 40.37\color{ForestGreen}\small{$(+5.67)$}\\
Diff-AUG~\cite{wang2024enhance}      & 49.02 & 40.59 &    48.71       &   38.58        &  44.23\color{ForestGreen}\small{$(+9.53)$}\\
Diff-Mix~\cite{wang2024enhance}      & 33.84            & 30.53            & 34.72 & 28.19 &  31.82\color{red}\small{$(-2.88)$}\\
\textbf{Ours} &  \cellcolor{mygray}\textbf{49.84}      &  \cellcolor{mygray}\textbf{41.46}         &  \cellcolor{mygray}\textbf{51.01}        &   \cellcolor{mygray}\textbf{42.03}       &  \cellcolor{mygray}\textbf{46.09}\color{ForestGreen}\small{$(+11.39)$}   \\ 
\hline
\hline
\end{tabular}%
}
\vspace{-0.7em}
\captionof{table}{\textbf{OOD classification.} ``L'', ``W'' represent ``land'' and ``water'', respectively. Results
are averaged on three trials.}
\vspace{-0.5em}
\label{table:3}
\end{figure}
\addtolength{\tabcolsep}{2pt}
\begin{figure}[t]
\centering
\resizebox{0.8\columnwidth}{!}{%
\def\arraystretch{0.92}
\begin{tabular}{ccccccc}
\hline\hline
\texttt{CL} & \multicolumn{1}{l}{\texttt{LI}} & \texttt{SI} & \texttt{SE} & \texttt{TD}  & LPIPS$(\uparrow)$  & Acc$(\uparrow)$   \\ \hline       &          &    &    &            &     49.9\%  &  31.53   \\
  \checkmark         &          &    &    &              &     47.9\%  & 38.74     \\\checkmark &
\checkmark                      &    &    &             & 27.5\%     & 37.32    \\\checkmark&
\checkmark                      &    &    & \checkmark             & 30.4\%     & 38.37     \\\checkmark&
                       & \checkmark  &    &          & 35.4\% & 37.70 \\\checkmark &
                       & \checkmark  & \checkmark  &          & 51.5\% & 40.49 \\ \checkmark&
                       &    &    & \checkmark        & 50.0\% & 40.78 \\\cellcolor{mygray}\checkmark &
                       \cellcolor{mygray}& \cellcolor{mygray}\checkmark  & \cellcolor{mygray}\checkmark  & \cellcolor{mygray}\cellcolor{mygray}\checkmark  &       \cellcolor{mygray}\textbf{52.7}\% & \cellcolor{mygray}\textbf{42.15} \\ \hline\hline
\end{tabular}%
}
\vspace{-0.7em}
\captionof{table}{\textbf{Components Ablation.}``\texttt{CL}'' is Concept Learning, ``\texttt{LI}'' is Linear Interpolation, ``\texttt{SI}'' is Spherical Interpolation, ``\texttt{SE}'' is Spherical Extrapolation and ``\texttt{TD}" is Two-stage Denosing.}
\label{table:4}
\end{figure}

\begin{figure}[t]
\centering
\vspace{-1.5em}
\includegraphics[width=0.85\linewidth]{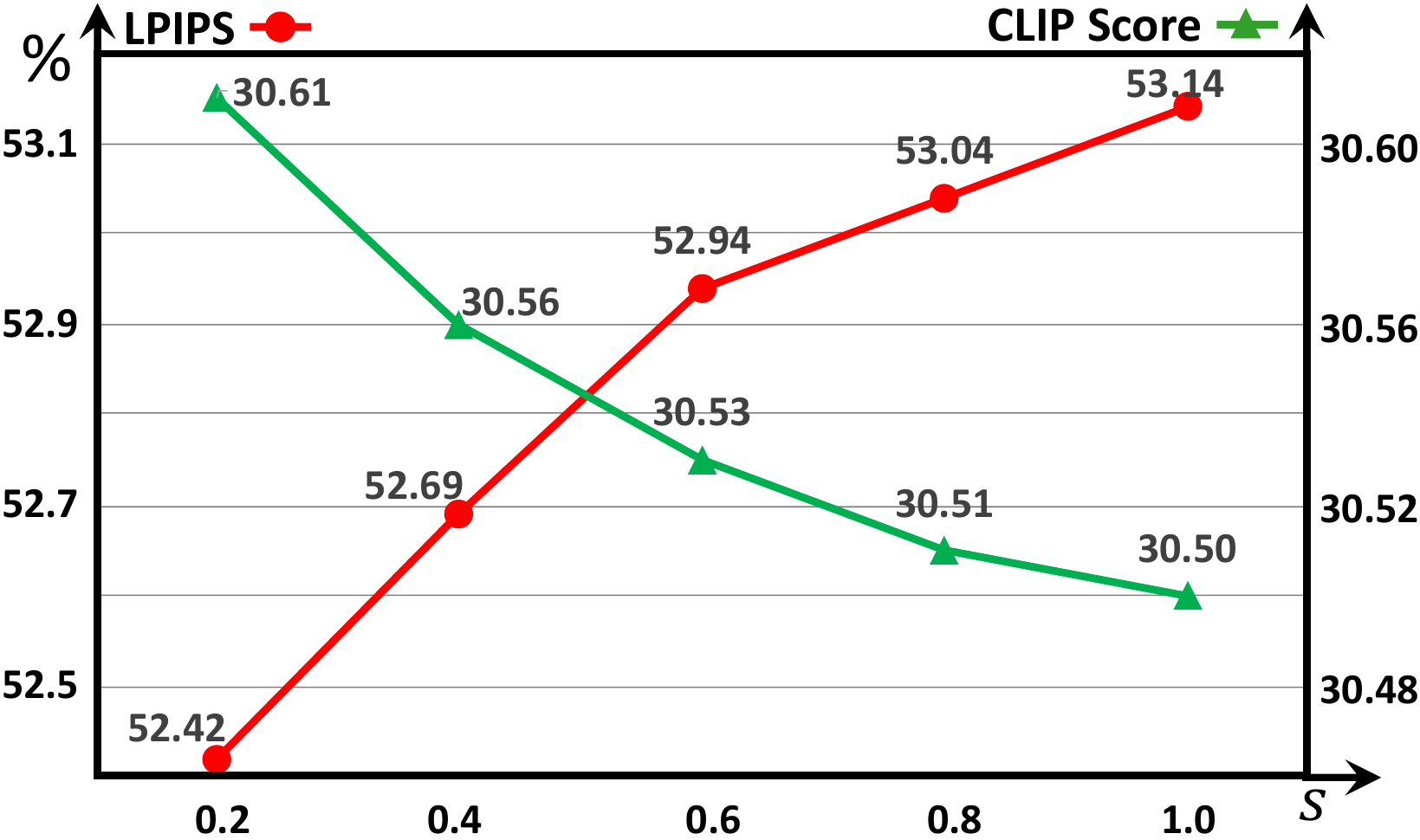}
\vspace{-0.5em}
\caption{\textbf{Influence of split ratio \textbf{$s$}}. Except for the split ratio, all other settings and hyperparameters are the same with 5-
shot CUB classification with ResNet-50.}
\label{fig:6}
\end{figure}

\noindent\textbf{Split Ratio.} Recall that in the two-stage denoising (\cf, Sec.~\ref{sec:3.3}), we have a split ratio $s$ to divide the denoising into two stages. To explore how the split ratio influences the synthetic data, we ablated it in Figure~\ref{fig:6}. This figure gives the curves of the \emph{CLIP score} of the synthetic set and average \emph{LPIPS} between images of the synthetic set changing with $s$. We can see that, with the increasing $s$, the \emph{CLIP score} decreases at a relatively slow rate while the \emph{LPIPS} has a relatively large increase. By adjusting $s$, a trade-off between faithfulness and diversity can be made.

\noindent\textbf{Qualitative Results.} 
In Figure~\ref{fig:7}, we give some visualizations of Da-Fusion~\citep{trabucco2023effective} and ours. We can see that the samples generated by DA-fusion lack diversity. In contrast, ours can generate samples with new context while maintaining the category characteristics.


\section{Conclusion}
\label{sec:5}

In this work, we analyze current diffusion-based DA methods from a unified perspective, finding that they can either only improve the faithfulness of synthetic samples or only improve their diversity. To take both faithfulness and diversity into account, we propose Diff-II, a simple yet effective diffusion-based DA method. Our Diff-II shows that it significantly improves both the faithfulness and diversity of the synthetic samples, further improving classification models in data-scarce sceneries. In the future, we are going to: 1) Extend this work into more general perception tasks, such as object detection, segmentation, or even video-domain tasks. 2) Explore how to remove the dependency of captioner and only leverage LLMs to diversify the prompts used in two-stage denoising, making the augmentation more robust and can better handle out-of-distribution test sets.

\noindent\textbf{Limitations.} 
Our method is less effective when some categories only have one training image. In that case, the interpolation can not be implemented because the interpolation operation is between two samples. We can see that for the long-tail classification task on CUB-LT (\cf, Table~\ref{table:2}): as the imbalance factor gets larger (from 10 to 100), the gain of our method (compared with the second best one Diff-Mix) is getting smaller and smaller (from $5.8\%$ to $0.86\%$). This is because a higher imbalance factor means there are more categories that only have one training image.

{
\small

\bibliographystyle{ieeenat_fullname}
}


\end{document}